\documentclass[12pt]{article}
\usepackage{hyperref}
\usepackage{amsmath}
\usepackage{amssymb}
\usepackage{comment}
\usepackage{algorithm}
\usepackage{algpseudocode}
\usepackage{graphicx,psfrag,epsf}
\usepackage{graphicx}
\usepackage{enumerate}
\usepackage{natbib}
\usepackage{amsfonts}
\usepackage{graphicx}
\usepackage{float}
\usepackage{booktabs}
\usepackage{tikz}
\usepackage{natbib}
\usepackage[figuresright]{rotating}
\usepackage{multirow}
\usepackage{bm}
\usepackage{listings}
\usepackage{url} 

\newtheorem{theorem}{Theorem}[section]  
\newtheorem{definition}[theorem]{Definition}  
\newtheorem{proof}{Proof}

\newcommand{\blind}{1}

\begin{document}

\def\spacingset#1{\renewcommand{\baselinestretch}%
{#1}\small\normalsize} \spacingset{1}
\if1\blind
{
  \title{\bf Fast Gibbs Sampling on Bayesian Hidden Markov Model with Missing Observations}
  \author{Li, Dongrong\hspace{.2cm}\\
    \small Department of Statistics and Data Science, The Chinese University of Hong Kong\\
    and \\
     Yu, Tianwei\hspace{.2cm}\\
    \small School Data Science, The Chinese University of Hong Kong, Shenzhen (CUHK-Shenzhen)\\
    and \\
     Fan, Xiaodan\thanks{
    Correspondence: xfan@cuhk.edu.hk} \\
    \small Department of Statistics and Data Science, The Chinese University of Hong Kong}
  \maketitle
} \fi

\if0\blind
{
  \bigskip
  \bigskip
  \bigskip
  \begin{center}
    {\LARGE\bf Title}
\end{center}
  \medskip
} \fi
\bigskip
\begin{abstract}
The Hidden Markov Model (HMM) is a widely-used statistical model for handling sequential data. However, the presence of missing observations in real-world datasets often complicates the application of the model. The EM algorithm and Gibbs samplers can be used to estimate the model, yet suffering from various problems including non-convexity, high computational complexity and slow mixing. In this paper, we propose a collapsed Gibbs sampler that efficiently samples from HMMs' posterior by integrating out both the missing observations and the corresponding latent states. The proposed sampler is fast due to its three advantages. First, it achieves an estimation accuracy that is comparable to existing methods. Second, it can produce a larger Effective Sample Size (ESS) per iteration, which can be justified theoretically and numerically. Third, when the number of missing entries is large, the sampler has a significant smaller computational complexity per iteration compared to other methods, thus is faster computationally. In summary, the proposed sampling algorithm is fast both computationally and theoretically and is particularly advantageous when there are a lot of missing entries. Finally, empirical evaluations based on numerical simulations and real data analysis demonstrate that the proposed algorithm consistently outperforms existing algorithms in terms of time complexity and sampling efficiency (measured in ESS).
\end{abstract}
\noindent%
\vfill

\newpage
\spacingset{1.45} 
\section{Introduction}
\label{sec:intro}
The Hidden Markov Model (HMM) is a powerful statistical tool for sequential data analysis. It assumes that we observe a series of outputs $\mathbf{y}$ that are determined by unobserved latent variables $\mathbf{z}$, which is a Markov chain. Since it can be considered as a way of modeling observations that come from a transformation or corruption of an unobserved latent sequence, it has found widespread applications in diverse fields related to sequence analysis such as DNA sequencing (\cite{hmm-dna}), text analysis (\cite{hmm-text}), speech recognition (\cite{Rabiner, bahl}), music analysis (\cite{hmm-music}), and medical health record analysis (\cite{hmm-ehr}). While extensive literature has addressed estimation and prediction problems in hidden Markov models (\cite{Rabiner, bishop2006pattern}), relatively few studies have focused on HMMs with missing observations, which commonly occur in medical and health time series such as Electronic Health Records (EHR) (\cite{hmm-ehr, yeh2012}), where missing entries originate from patients' irregular visits or hospitals' irregular recordings.

Although several approaches have been proposed to tackle this problem (as we shall introduce later), they generally suffer from high computational complexity. In this paper, we propose a computationally efficient Gibbs sampling method for posterior inference in hidden Markov models with missing observations. The proposed sampler employs a novel forward-backward algorithm that analytically integrates out both the missing data and their corresponding latent states, thereby improving the efficiency of posterior sampling in terms of theoretical convergence rate and time complexity. Furthermore, we demonstrate that the proposed algorithm exhibits particularly strong advantages in high missing-rate scenarios due to its significantly faster convergence under such conditions.

To provide a foundation for our subsequent discussion, we first review related methods and approaches in this area. Markov modeling with incomplete sequences was originally introduced in the seminal works of \cite{deltour1999}, \cite{albert2000}, and \cite{Yeh2010}. However, these methods cannot be directly applied to hidden Markov models. Since the 21st century, various approaches have emerged for modeling incomplete sequences with HMM-like models.

Hidden semi-Markov models with categorical outcomes were investigated in \cite{hidden-semi} to address the missing data problem. In \cite{hmm-test}, different missing mechanisms in hidden Markov models were systematically studied, with a statistical test proposed to distinguish between these mechanisms. \cite{cooke2001} explored the estimation of HMMs with incomplete sequences using simple imputation strategies such as mode imputation. However, this approach remains relatively ad-hoc, as the performance of mode/forward imputation in HMMs with missing data lacks theoretical guarantees.

The problem of imputing missing values in Factorial Hidden Markov Models (\cite{fhmm}) was addressed more systematically in \cite{lee2008}. Nevertheless, this work primarily focuses on imputing missing observations under a model with known true parameters.

The problem of hidden Markov models with missing observations was formally investigated in \cite{yeh2012} and \cite{Maarten2021}. In \cite{yeh2012}, an EM algorithm was proposed for parameter estimation, with its performance evaluated through simulation experiments under different missing mechanisms. \cite{Maarten2021} examined several non-ignorable missing mechanisms (\cite{Rubin2019}) and discussed parameter estimation under these frameworks.

Both studies operate on a likelihood function with missing observations analytically integrated out, employing EM algorithms (\cite{EM,dempster-em}) for parameter estimation through iterative updates based on forward-backward probability calculations (\cite{Rabiner}). As both EM algorithms and Gibbs sampling constitute coordinate descent methods (\cite{em-vs-mcmc}), a Gibbs sampler (\cite{mcmc,gibbs}) targeting the same objective function can be derived as a randomized variant of the approaches presented in \cite{hidden-semi,yeh2012,Maarten2021}. This randomized formulation demonstrates reduced susceptibility to local minima or flat regions in the parameter space (\cite{em-vs-mcmc}). A comprehensive description of Bayesian Gibbs sampling for hidden Markov models can be found in \cite{hmm-em-vs-mcmc}.

All research works described above employ iterative algorithms that rely on parameter updates through forward-backward probability calculations, resulting in a computational complexity of $O(nT)$ for $n$ sequences with average length $T$. As this paper demonstrates, we propose a novel Gibbs sampling algorithm that significantly reduces this computational time complexity as the number of missing entries increases, thereby enabling more efficient posterior estimation. The proposed sampler integrates out extraneous latent states, consequently operating in a reduced parameter space and achieving faster theoretical convergence rates.

Beyond discrete-time categorical HMMs, substantial research has been dedicated to continuous-time HMMs for modeling partially-observed sequences (\cite{liu2015efficient,continuous-hmm,time-discrete}). However, the application of continuous-time HMMs to discrete-time estimation scenarios faces inherent limitations. This challenge stems from the requirement to estimate sojourn times in continuous-time frameworks – a process that introduces additional uncertainty in both parameter estimation and state prediction when analyzing discrete time series.

Based on the existing literature, we summarize our contributions as follows:

First, we propose an accelerated Gibbs sampler through analytical marginalization of missing observations and their corresponding latent states. We introduce a novel collapsed posterior sampling methodology based on an optimized forward-backward algorithm. Second, we establish the computational efficiency of the proposed algorithm through rigorous complexity analysis and convergence rate characterization. Furthermore, we demonstrate that this approach exhibits particular advantages in high-missingness regimes, achieving significantly faster runtime compared to existing methods when processing datasets with large proportions of missing observations. Finally, we validate these claims through comprehensive numerical experiments, showing superior performance in both computational speed and Effective Sample Size (ESS).

The remainder of this paper is organized as follows: We first review fundamental concepts of hidden Markov models and missing data mechanisms. We then present the proposed algorithm in detail. We then provides theoretical analyses of computational complexity and convergence rates, highlighting comparative advantages over existing methods. Finally, we demonstrate the algorithm's effectiveness through simulation studies and real-world data experiments, with quantitative comparisons of runtime efficiency and sampling performance.

\subsection{Hidden Markov Models}
\label{sec:intro_hmm}
Let $\mathbf{y}=(y_1,...,y_T)$ be an observed sequence with length $T$. A hidden Markov model with parameters $\bm{\theta}$ is a generative model that assumes $\mathbf{y}$ is generated by the following procedure:
$$
p(\mathbf{y},\mathbf{z}|\bm{\theta})=p(\mathbf{z}|\bm{\theta})p(\mathbf{y}|\mathbf{z},\bm{\theta}),
$$
where $\mathbf{z}$ is a sequence of unobserved latent states. In the HMM framework, $p(\mathbf{z}|\bm{\theta})$ is typically defined as a Markov chain:
$$
p(\mathbf{z}|\bm{\theta})=p(z_1|\bm{\theta})\prod_{t=1}^{T-1}p(z_{t+1}|z_t,\bm{\theta}),
$$
with conditionally independent observations given the latent states:
$$
p(\mathbf{y}|\mathbf{z},\bm{\theta})=\prod_{t=1}^T p(y_t|z_t,\bm{\theta}).
$$
When $z_t$ and $y_t$ are defined over finite state spaces, the standard parameterization is $\bm{\theta}=(\bm{\pi},\mathbf{A},\mathbf{B})$, where $\bm{\pi}$ denotes the initial state distribution, $\mathbf{A}$ is the state transition matrix, and $\mathbf{B}$ contains emission probabilities with $B_{ij}=P(y=j|z=i)$ for valid states $i$ and $j$.

From a Bayesian perspective, the hierarchical formulation becomes:
$$
p(\mathbf{y},\mathbf{z},\bm{\theta})=p(\bm{\theta})p(\mathbf{z}|\bm{\theta})p(\mathbf{y}|\mathbf{z},\bm{\theta}),
$$
where $p(\bm{\theta})$ represents prior distributions for the model parameters.

In the context of hidden Markov modeling, our primary interest lies in estimating both model parameters and latent sequences (\cite{Rabiner}). A common approach involves treating $\mathbf{z}$ as a latent variable and applying the EM algorithm (\cite{dempster-em,EM}) to obtain maximum likelihood estimates. For latent sequence estimation, MAP (\cite{bishop2006pattern}) or marginalized MAP estimators are typically employed. The Viterbi algorithm (\cite{forney1973viterbi}) provides an effective dynamic programming solution for sequence-level MAP estimation (\cite{bellman1966dynamic}). Alternatively, Gibbs sampling with data augmentation (\cite{liu1999da}) enables joint estimation of parameters and latent states, with detailed implementations described in \cite{hmm-em-vs-mcmc}. Both approaches exhibit similar theoretical performance since Gibbs sampling can be interpreted as a randomized coordinate descent variant (\cite{em-vs-mcmc}).

Building on these foundations, MCMC methods based on Gibbs sampling (\cite{hmm-em-vs-mcmc}) have been developed, which alternate between parameter and latent state updates as a randomized coordinate descent procedure. Empirical evidence suggests MCMC algorithms can escape local modes, demonstrating superior convergence behavior compared to EM (\cite{hmm-em-vs-mcmc}). Additional computational approaches include variational inference (\cite{blei2017,hmm-vi}), though this method approximates the posterior through variational distributions that may introduce estimation bias and instability.

All these aforementioned algorithms require computation of forward probabilities $\alpha_t(i) = P(y_1,\ldots,y_t,z_t=i\mid\bm{\theta})$ and backward probabilities $\beta_t(i){=}P(y_{t+1},\ldots,y_T\mid z_t{=}i,\bm{\theta})$ for $t{=}1,\ldots,T$ to perform the iterative updates. As these probabilities lack closed-form solutions, they must be computed recursively through dynamic programming, typically requiring $O(T)$ operations per iteration. \cite{Rabiner} provides comprehensive details on their recursive computation and application to parameter updates.

In the forthcoming sections, we propose a collapsed Gibbs sampler. Under this new sampling scheme, a new strategy is adopted to evaluate the forward and backward probabilities, which is significantly faster than existing algorithms when there are missing observations.

\subsection{Notations}
In the following section, we commence our formal analysis. To facilitate understanding, we provide a concise overview of the notations frequently employed throughout this paper.

Individual numbers or elements originating from a specific set are represented by lowercase letters, such as $z$ or $y$. A vector or A sequence of the elements are denoted by bold face English letters such as $\mathbf{z}$ or $\mathbf{y}$. $\mathbf{z}_{t_1:t_2}$ is used to represent values from position $t_1$ to position $t_2$ (both sides included) in a sequence $\mathbf{z}$. We also use Greek letters such as $\pi$ to denote vectors or a collection of vectors. We emphasize that $\theta$ denotes the collection of all parameters. When it comes to matrices, or collections thereof, we utilize boldface letters like $\mathbf{A}$ or $\mathbf{B}$. The notation $\bm{\Delta}^{k}$ signifies a vector defined within a $(k-1)$-dimensional simplex, while $\bm{\Delta}^{r\times k}$ designates an $r\times k$ matrix, with each row being a vector defined on a $(k-1)$-dimensional simplex.

Sets are represented using script fonts or uppercase letters (excluding $\bm{\Delta}$), such as $\mathcal{Z}, \mathcal{Y}$, and $\Omega$. The cardinality of a set $\mathcal{Z}$ is expressed as $|\mathcal{Z}|$. We use $\Omega$ to denote a sample space, and $P$ or $\mu$ to signify a particular probability measure. Consequently, the integration of a function $f$ according to the measure $\mu$ is represented as $\int_{\Omega}f(x)d\mu(x)$.

We use $T$ to denote the length of a sequence and let $n$ denote the sample size of the entire dataset. We use $p$ to denote the missing probability (as we shall define later).

\section{Method}
\label{sec:meth}
\subsection{Problem Formulation}
\label{sec: meth-pf}
In this section, we formalize the problem of Bayesian estimation for hidden Markov models with incomplete sequences. We systematically review existing methodologies and propose a novel algorithm demonstrating superior computational efficiency.

Let $\mathbf{y}$ represent the theoretical complete observation sequence. In our framework, we assume $\mathbf{y}$ is only partially observed, such that $\mathbf{y}=(\mathbf{y}_o,\mathbf{y}_m)$, where $\mathbf{y}_o$ denotes the observed component and $\mathbf{y}_m$ the missing component. Throughout our analysis, we maintain the ignorable missingness assumption (\cite{Rubin1976}).

Under this formulation, the joint distribution factorizes as:
$$
p(\bm{\theta},\mathbf{y},\mathbf{z})=p(\bm{\theta})p(\mathbf{z}|\bm{\theta})p(\mathbf{y}|\mathbf{z},\bm{\theta})=p(\bm{\theta})p(\mathbf{z}|\bm{\theta})p(\mathbf{y}_o|\mathbf{z},\bm{\theta})p(\mathbf{y}_m|\mathbf{z},\bm{\theta}),
$$
where the final equality follows from the conditional independence structure inherent in HMMs.

A direct approach to posterior inference involves data augmentation (\cite{gibbs}), requiring iterative sampling from the following conditional distributions (\cite{hmm-em-vs-mcmc}):
\begin{equation}
\begin{aligned}
    \mathbf{y}_m &| \mathbf{z}, \bm{\theta}, \\
    \bm{\theta} &| \mathbf{y}_o, \mathbf{z}, \\
    \mathbf{z} &| \bm{\theta}, \mathbf{y}_o.
\end{aligned}
\label{eq:conditional}
\end{equation}
This methodology, however, proves computationally inefficient. Prior work has developed an alternative approach by encoding missing values $\mathbf{y}_m$ as special tokens with emission probability fixed at 1 (\cite{yeh2012,Maarten2021}). This formulation is mathematically equivalent to analytically marginalizing $\mathbf{y}_m$ from the complete-data likelihood.

Under this framework, recursive computation becomes feasible for forward probabilities $\alpha_t(i) = P(\mathbf{y}_{1:t}, z_t = i \mid \bm{\theta})$ and backward probabilities $\beta_t(i) = P(\mathbf{y}_{t+1:T} \mid z_t = i, \bm{\theta})$, which drive iterative parameter updates. Detailed derivations of these probability recursions and their corresponding EM algorithm implementation have been rigorously established in prior literature (\cite{yeh2012}).

Given that both EM algorithms and Gibbs sampling constitute coordinate descent variants, we derive a corresponding Gibbs sampler demonstrating comparable performance. This approach involves iterative sampling from the following conditional distributions targeting the integrated joint distribution:

\begin{equation}
\begin{aligned}
    \bm{\theta} &| \mathbf{y}_o, \mathbf{z}, \\
    \mathbf{z} &| \bm{\theta}, \mathbf{y}_o.
\end{aligned}
\label{eq:ym-integrated}
\end{equation}

Sampling from the latent state conditional distribution $\mathbf{z}|\bm{\theta}, \mathbf{y}_o$ requires recursive evaluation of forward-backward probabilities. The complete computational procedure is documented in \cite{hmm-em-vs-mcmc}.

However, it is not difficult to notice that the above method requires evaluating the forward and backward probabilities for each position $t$, suggesting a computational complexity of $O(nT)$ per update (where $n$ denotes the sample size). When the sequence is long, this algorithm becomes slow. Additionally, define the missing rate (or missing probability) as
\begin{equation}
\label{eq:missing-rate}
p = \frac{|\mathbf{y}_m|}{|\mathbf{y}_m| + |\mathbf{y}_o|},
\end{equation}
representing the fraction of missing observations. When the missing rate is very high, the algorithm becomes unnecessarily complicated. In response, we propose a posterior sampling algorithm with a time complexity of $O((1-p)nT)$ per iteration. This means that the proposed algorithm will be significantly faster than that described in \ref{eq:conditional} when the missing rate $p$ is high. Furthermore, \ref{eq:conditional} also has a larger exploration space (compared to the proposed method), which will slow down the convergence speed in Gibbs sampling. As we will argue in the forthcoming sections, this parameter space can be significantly reduced in the presence of missing data, which can improve the convergence speed. Therefore, our proposed algorithm is advantageous not only in terms of computational complexity but also in convergence rate.

\subsection{Description of the Collapsed Model}
In this section, we derive the collapsed joint distribution through analytical marginalization of redundant latent states, with posterior sampling techniques discussed subsequently.

Our derivation begins with the complete-data likelihood:
$$
p(\bm{\theta},\mathbf{y},\mathbf{z}) = p(\bm{\theta})p(\mathbf{z}|\bm{\theta})p(\mathbf{y}_o|\bm{\theta},\mathbf{z})p(\mathbf{y}_m|\bm{\theta},\mathbf{z}),
$$
where the equality follows from the conditional independence structure.

Through integration over missing observations, we obtain:
$$
p(\bm{\theta},\mathbf{y}_m,\mathbf{z}) = \int p(\bm{\theta})p(\mathbf{z}|\bm{\theta})p(\mathbf{y}_o|\bm{\theta},\mathbf{z})p(\mathbf{y}_m|\bm{\theta},\mathbf{z})d\mu(\mathbf{y}_m).
$$
This formulation underpins the methodology in \cite{yeh2012,Maarten2021}.

Crucially, we demonstrate that further complexity reduction can be achieved by decomposing the latent states into $\mathbf{z} = (\mathbf{z}_m, \mathbf{z}_o)$, where $\mathbf{z}_m$ corresponds to missing observations $\mathbf{y}_m$ and $\mathbf{z}_o$ corresponds to observed data $\mathbf{y}_o$.

This yields the refined joint distribution:
$$
p(\bm{\theta}, \mathbf{y}_o, \mathbf{z}) = p(\bm{\theta}) \, p(\mathbf{z}_m, \mathbf{z}_o | \bm{\theta}) \, p(\mathbf{y}_o | \mathbf{z}_o, \bm{\theta}).
$$

Therefore, we can further integrate the latent states that corresponds to the missing observations (i.e. $\mathbf{z}_m$ out:
$$
\begin{aligned}
    p(\bm{\theta}, \mathbf{y}_o, \mathbf{z}_o) &= \int p(\bm{\theta}, \mathbf{y}_o, \mathbf{z}_o, \mathbf{z}_m) \, d\mu(\mathbf{z}_m)\\
    &= p(\bm{\theta}) \int p(\mathbf{z}_o, \mathbf{z}_m | \bm{\theta}) \, p(\mathbf{y}_o | \mathbf{z}_o, \bm{\theta}) \, d\mu(\mathbf{z}_m)\\
    &= p(\bm{\theta}) \, p(\mathbf{y}_o | \mathbf{z}_o, \bm{\theta}) \int p(\mathbf{z}_o, \mathbf{z}_m | \bm{\theta}) \, d\mu(\mathbf{z}_m).
\end{aligned}
$$

Without loss of generality, assume that $y_1$ is observed and therefore $z_1\in \mathbf{z}_o$. A further calculation shows:
$$
\begin{aligned}
    \int p(\mathbf{z}_o,\mathbf{z}_m|\bm{\theta}) d\mu(\mathbf{z}_m) &= \int p(z_1|\bm{\theta}) [\prod_{t=1}^{T-1} p(z_t|z_{t-1},\bm{\theta})] d\mu(\mathbf{z}_m)\\
    &= \int p(z_1|\bm{\theta}) [\prod_{t=1}^{T-1} p(z_t|z_{t-1},\bm{\theta})] dz_{t_1}...dz_{t_k}\\
    &= p(z_1|\bm{\theta}) \prod_{t_i \in o} p(z_{t_{i+1}}|z_{t_i},\bm{\theta})\\
    &= p(z_1|\bm{\theta}) \prod_{t_i \in o}  \mathbf{1}^T_{z_{t_{i+1}}} \mathbf{A}^{t_{i+1}-t_i}\mathbf{1}_{z_{t_i}},
\end{aligned}
$$
where $z_{t_k}$ stands for the latent states in $\mathbf{z}_m$, $o$ stands for the index that corresponds to the observed states and $\mathbf{A}$ stands for the transition matrix. The above calculation shows that the latent states corresponding to the missing observations are in fact redundant and can be integrated out analytically.

This collapsed distribution enables a Gibbs sampler with iterative updates from the following conditional distributions:
\begin{equation}
\begin{aligned}
    \bm{\theta} &| \mathbf{y}_o, \mathbf{z}_o, \\
    \mathbf{z}_o &| \bm{\theta}, \mathbf{y}_o.
\end{aligned}
\label{eq:collapsed}
\end{equation}

These conditional distributions reduce the parameter space dimensionality of latent states, thereby accelerating convergence rates (as detailed in subsequent sections). Furthermore, both forward/backward probability calculations and $\mathbf{z}_o$ sampling achieve improved time complexity -- particularly advantageous in high missing probability regimes ($p \to 1$).

\subsection{Posterior Sampling}
\label{sec:post-sampling}
In this section, we develop a Gibbs sampler targeting the collapsed distribution $p(\bm{\theta}, \mathbf{y}_o, \mathbf{z}_o)$. Our discussion focuses particularly on $\mathbf{z}_o$ state sampling, which demonstrates superior time complexity compared to conventional approaches.

\subsubsection{Sampling from the Conditional of $\mathbf{z}_o$}
\label{sec:sample-chain}
First, we note that $\mathbf{z}_o$ forms an inhomogeneous Markov chain, where the transition probability between $z_{t_k}$ and $z_{t_{k+1}}$ (for $z_{t_k}, z_{t_{k+1}} \in \mathbf{z}_o$) is parameterized as $\mathbf{A}^{t_{k+1} - t_k}$. The forward probability $\alpha_{t_k}(z_{t_k}) := p((\mathbf{y}_o)_{1:t_k},z_{t_k}|\bm{\theta})$ can be recursively computed as:
\begin{equation}
\label{eq:forward-prob}
\begin{aligned}
    \alpha_{t_k}(z_{t_k}) &= p((\mathbf{y}_o)_{1:t_k},z_{t_k}|\bm{\theta})\\
    &= \sum_{z_{t_{k-1}}} p(z_{t_k}, z_{t_{k-1}},(\mathbf{y}_o)_{1:t_{k}-1}|\bm{\theta})\\
    &= \sum_{z_{t_{k-1}}} p(z_{t_k}, z_{t_{k-1}},(\mathbf{y}_o)_{1:t_k}|\bm{\theta})\\
    &= p(y_{t_k}|z_{t_k},\bm{\theta}) \sum_{z_{t_{k-1}}} \alpha_{t_{k-1}}(z_{t_{k-1}}) p(z_{t_k}|z_{t_{k-1}},\bm{\theta})\\
    &= p(y_{t_k}|z_{t_k},\bm{\theta}) \sum_{z_{t_{k-1}}} \alpha_{t_{k-1}}(z_{t_{k-1}}) \mathbf{1}^T_{z_{t_{k}}}\mathbf{A}^{t_{k} - t_{k-1}}\mathbf{1}_{z_{t_{k-1}}}.
\end{aligned}
\end{equation}

Computing forward probabilities for all states in $\mathbf{z}_o$ requires $O((1-p)NT)$ operations with $N$ chains of length $T$ and missing rate $p$. Although matrix exponential operations are involved, they can be efficiently executed using numerical linear algebra libraries. Furthermore, the results of these matrix exponential can be precomputed and cached. Consequently, each subsequent matrix exponential evaluation incurs only $O(1)$ time complexity through cached results. In practice, when the sample size $N$ is large, the computational overhead from matrix operations becomes negligible.

Evaluating the forward probability helps us to sample the latent states $\mathbf{z}_o$ backwardly. Let $o = \{t_1,..., t_K\}$, the conditional distribution of $\mathbf{z}_o$ has the following backward decomposition:
$$
\begin{aligned}
    p(\mathbf{z}_o|\mathbf{y}_o,\bm{\theta})&= p(z_{t_K}|\mathbf{y}_o,\bm{\theta})\prod_{t_k \in o, k\ne K}p(z_{t_k}|\mathbf{z}_{t_{k+1}:t_K},\mathbf{y}_o,\bm{\theta}).\\
\end{aligned}
$$
For the first term, it is easy to see that $p(z_{t_K}|\mathbf{y}_o,\bm{\theta}) \propto \alpha_{t_k}(z_{t_k})$.

For the second term, the following decomposition holds:

\begin{equation}
\label{eq:backward-prob}
\begin{aligned}
    p(z_{t_k}|\mathbf{z}_{t_{k+1}:t_K},\mathbf{y}_o,\bm{\theta})&= p(z_{t_k}|z_{t_{k+1}}, (\mathbf{y}_o)_{1:t_k}, (\mathbf{y}_o)_{t_{k+1}:t_K},\bm{\theta})\\
    &= p(z_{t_k}|z_{t_{k+1}},(\mathbf{y}_o)_{1:t_k},\bm{\theta})\\
    &\propto \frac{p(z_{t_{k+1}}|z_{t_k},\bm{\theta}) p(z_{t_k}|\mathbf{y}_o)_{1:t_k},\bm{\theta})}{\sum_{z\in\mathcal{Z}}p(z_{t_{k+1}}|z_{t_k}=z,\bm{\theta})p(z_{t_k}=z|\mathbf{y}_o)_{1:t_k},\bm{\theta})}\\
    &=\frac{p(z_{t_{k+1}}|z_{t_k},\bm{\theta}) p(z_{t_k}|\mathbf{y}_o)_{1:t_k},\bm{\theta})}{\sum_{z\in\mathcal{Z}}p(z_{t_{k+1}}|z_{t_k}=z,\bm{\theta})\alpha_{t_k}(z)}\\
    &=\frac{(\mathbf{1}^T_{z_{t_{k+1}}}\mathbf{A}^{t_{k+1} - t_k}\mathbf{1}_{z_{t_k}}) \alpha_{t_k}(z_{t_k})}{\sum_{z\in\mathcal{Z}}(\mathbf{1}^T_{z_{t_{k+1}}}\mathbf{A}^{t_{k+1} - t_k}\mathbf{1}_{z})
    \alpha_{t_k}(z)},
\end{aligned}
\end{equation}

which provides us an explicit way to evaluate the conditional probability for each latent state in the observed index $o$. With this decomposition, we can first forwardly compute $\bm{\alpha}_{t_k}$ for $t_k \in o$ and then backwardly sample $z_{t_K}, z_{t_{K-1}},..., z_{t_1}$ sequentially. Similar to the forward recursion procedure, the backward sampling approach also has a time complexity of $O((1-p)NT)$ and the computation cost on matrix multiplication is negligible. A full description of the sampling algorithm can be found in Algorithm~\ref{ffbs-alg}.

\begin{algorithm}
\caption{Forward-Backward Sampling of the Collapsed Gibbs Sampler}
\label{ffbs-alg}
\begin{algorithmic}[1]
\State \textbf{Input:} Observations $\mathbf{y}_o$, transition matrix $\mathbf{A}$, other parameters $\bm{\theta}$(including emission distribution and initial probability)
\State \textbf{Output:} Sample of hidden states $\mathbf{z}_o$

\\ \textit{Forward Recursion}
\For{$k = 1$ to $K$}
    \For{each state $z_t$}
        \State $\alpha_t(z_t) = p(y_{t_k}|z_{t_k},\bm{\theta}) \sum_{z_{t_{k-1}}} \alpha_{t_{k-1}}(z_{t_{k-1}}) \mathbf{1}^T_{z_{t_{k+1}}}\mathbf{A}^{t_{k+1} - t_k}\mathbf{1}_{z_{t_{k}}}$
    \EndFor
\EndFor

\\ \textit{Backward Sampling}
\For{$t = K$ down to $1$}
    \State $p(z_{t_k}|\mathbf{z}_{t_{k+1}:t_K},\mathbf{y}_o,\bm{\theta}) \propto \frac{(\mathbf{1}^T_{z_{t_{k+1}}}\mathbf{A}^{t_{k+1} - t_k}\mathbf{1}_{z}) \alpha_{t_k}(z_{t_k})}{\sum_{z\in\mathcal{Z}}(\mathbf{1}^T_{z_{t_{k+1}}}\mathbf{A}^{t_{k+1} - t_k}\mathbf{1}_{z})
    \alpha_{t_k}(z)} $
    \State Sample $z_{t_k}$ from $p(z_{t_k}|\mathbf{z}_{t_{k+1}:t_K},\mathbf{y}_o,\bm{\theta})$
\EndFor

\State \Return $\mathbf{z}_{1:T}$
\end{algorithmic}
\end{algorithm}

\subsubsection{Sampling From the Conditional of the Parameters}
\label{sec:sample-params}
In this section, we discuss the issue of sampling from the conditional of the parameters, including the transition matrix $\mathbf{A}$, emission matrix $\mathbf{B}$ and the initial distribution $\bm{\pi}$.

For the emission matrix $\mathbf{B}$, we can sample its $i$-th row one after another. If $\mathbf{B}$ has a Dirichlet prior $Dir(\bm{\eta})$, then its conditional distribution is given by:
$$
\mathbf{B}_i|\mathbf{B}_{-i},\mathbf{y}_o,\mathbf{z}_o, \mathbf{A},\bm{\pi} \sim Dir(\eta_{ij} + n_{ij}),
$$
where $n_{ij}$ stands for the number of observations with hidden state $i$ and observed state $j$.

For parameters $\mathbf{A}$ and $\bm{\pi}$, their conditional distributions are not directly available analytically. However, the Metropolis-within-Gibbs scheme (\cite{mh-within-gibbs}) can be adopted to update the parameters, where the proposal can be chosen as the random walk (\cite{rwm}) or gradient-based updates (\cite{sgld-simplex}).

Finally, Algorithm~\ref{alg:gibbs} summarizes the complete collapsed Gibbs sampler, which alternately samples parameters and latent sequences $\mathbf{z}_o$ via the forward-backward procedure in Algorithm~\ref{ffbs-alg}.

\begin{algorithm}
\small
\caption{Collapsed Gibbs Sampling for HMMs with Incomplete Observations}
\label{alg:phmmS}
\label{alg:gibbs}
\begin{algorithmic}
\Require Initial estimates $\bm{\theta}_0$ of the parameters $\bm{\theta}$, number of draws from the posterior $N$
\State $i \gets 1$
\While{$i\leq N$}
\State Sample $\bm{\theta}_k$ from conditional distribution $p(\bm{\theta}|\mathbf{z}_o,\mathbf{y}_o)$
\State $k \gets 1$
\While{$k \leq n$}
\State Sample the latent sequence $(\mathbf{z}_k)_o$ with Algorithm~\ref{ffbs-alg}
\State $k \gets k+1$
\EndWhile
\State $i \gets i+1$
\EndWhile
\State Discard the samples produced in the burn-in period and keep the remained samples as draws from the posterior.
\end{algorithmic}
\end{algorithm}

\subsection{Predictive Distribution}
\label{sec: meth-pred}
The proposed methodology's requirement of marginalized latent states necessitates efficient predictive procedures for latent states and missing values. This section develops computational frameworks to obtain predictive distributions under our sampling scheme through three primary objectives: forecasting future states, reconstructing latent sequences, and imputing missing observations. Each predictive task requires explicit sampling of latent trajectories $\mathbf{z}$ via forward-backward recursions.

\paragraph{Forecasting} Notice that $\mathbf{z}_k$ is itself a Markov chain with length $T$. We can certainly discuss its future behavior in the time window $T+1 : T+W$. For instance, in medical record analyses, it is of interest to predict the future behavior and progression pattern of a patient's health status based on his or her incomplete observation sequence $\mathbf{y}_o$. Assume we are interested in drawing the sequence $\tilde{\mathbf{z}} = (z_{1}, \dots, z_{T}, z_{T+1}, \dots, z_{T+W}) = (\mathbf{z}, \mathbf{z}^W)$. The predictive distribution of $\tilde{\mathbf{z}}$ is then given by the following computation:

$$
\begin{aligned}
p(\tilde{\mathbf{z}}|\mathbf{y}_o) &= p(\mathbf{z}, \mathbf{z}^W | \mathbf{y}_o) \\
&= \int_{\Omega} p(\mathbf{z}, \mathbf{z}^W | \bm{\theta}, \mathbf{y}_o) p(\bm{\theta} | \mathbf{y}_o) d\mu(\bm{\theta}) \\
&= \int_{\Omega} p(\mathbf{z} | \bm{\theta}, \mathbf{y}_o) p(\mathbf{z}^W | \bm{\theta}, \mathbf{z}, \mathbf{y}_o) p(\bm{\theta} | \mathbf{y}_o) d\mu(\bm{\theta}) \\
&= \int_{\Omega} p(\mathbf{z} | \bm{\theta}, \mathbf{y}_o) p(\mathbf{z}^W | \bm{\theta}, \mathbf{z}) p(\bm{\theta} | \mathbf{y}_o) d\mu(\bm{\theta}).
\end{aligned}
$$

Computationally, we first draw $\bm{\theta}$ from its marginal posterior and the corresponding $\mathbf{z}$. For each drawn parameter and latent sequence, we let the sequence progress for another $W$ steps. Such a construction draws a sample from $p(\tilde{\mathbf{z}}|\mathbf{y}_o)$ and hence provides a valid predictive forecasting distribution. Sampling from $\mathbf{z}$ can be derived from the forward-backward probabilities given in \cite{yeh2012} with a procedure similar to the algorithm defined in Algorithm~\ref{ffbs-alg} or \cite{hmm-em-vs-mcmc}, which has a time complexity of $O(T)$.

latex

Copy
\paragraph{Predictive Distribution from a New Sample}
Another type of predictive distribution that is more common in Bayesian modeling involves predicting the corresponding latent sequence $\tilde{\mathbf{z}}$ given a new sequence $\tilde{\mathbf{y}}_o$. Notice that
$$
p(\tilde{\mathbf{z}}|\tilde{\mathbf{y}}_o,\mathbf{y}_o) \propto \int_{\Omega} p(\tilde{\mathbf{z}},\tilde{\mathbf{y}}_o|\bm{\theta}) p(\bm{\theta}|\mathbf{y}_o) d\mu(\bm{\theta}),
$$
which suggests that, computationally, providing a predictive distribution for $\tilde{\mathbf{z}}$ requires only drawing $\bm{\theta}$ from its marginal posterior and sampling each latent sequence from $\tilde{\mathbf{y}}_o$. As discussed in the previous section, this procedure also has a time complexity of $O(T)$.

\paragraph{Missing Observations Imputation}
In the context of Bayesian modeling, it is also of interest to impute the missing observations $\mathbf{y}_m$. The predictive distribution of $\mathbf{y}_m$ can be obtained via the following steps:
$$
\begin{aligned}
p(\mathbf{y}_m|\mathbf{y}_o) &= \int_{\Omega} p(\mathbf{y}_m, \mathbf{z}, \bm{\theta} | \mathbf{y}_o) d\mu(\bm{\theta}) d\mu(\mathbf{z}) \\
&= \int_{\Omega} p(\mathbf{y}_m | \mathbf{z}, \bm{\theta}, \mathbf{y}_o) p(\mathbf{z}, \bm{\theta} | \mathbf{y}_o) d\mu(\bm{\theta}) d\mu(\mathbf{z}) \\
&= \int_{\Omega} p(\mathbf{y}_m | \mathbf{z}, \bm{\theta}) p(\mathbf{z}, \bm{\theta} | \mathbf{y}_o) d\mu(\bm{\theta}) d\mu(\mathbf{z}).
\end{aligned}
$$
Computationally, we first draw $\bm{\theta}$ and $\mathbf{z}$ from their posterior distributions and sample $\mathbf{y}_m$ from $p(\mathbf{y}_m|\bm{\theta},\mathbf{z})$, which is specified by the emission distribution.

We emphasize that although the proposed method generally has a computational complexity of $O(n(1-p)T)$, sampling from its predictive posterior distribution on new sequences does have a time complexity that is proportional to $T$. Therefore, the advantages of the proposed algorithm lie mainly in its lower computational complexity and convergence rate (as discussed in Section~\ref{sec:complexity-analysis}) during training. In fact, the proposed algorithm is fast because it avoids the unnecessary prediction and imputation of $\mathbf{y}_m$ and $\mathbf{z}_o$ during the training procedure, which accelerates its convergence speed and reduces its time complexity.

\section{Complexity Analysis}
\label{sec:complexity-analysis}
In this section, we analyze the complexity of the proposed sampler. We argue that the proposed collapsed sampler is faster than existing methods in terms of convergence rate and computational complexity.

\subsubsection{Convergence Rate Analysis}
\label{sec:convergence-rate}
For fairness consideration, we mainly compare the convergence speed of the samplers described in Eq~\ref{eq:conditional}, Eq~\ref{eq:ym-integrated}, and Eq~\ref{eq:collapsed}. Let $\mathbf{F}_g$ be the transition kernel induced by the algorithm described in Eq~\ref{eq:conditional}, $\mathbf{F}_m$ be the transition kernel induced by the algorithm described in Eq~\ref{eq:ym-integrated}, and $\mathbf{F}_c$ be the transition kernel induced by the algorithm described in Eq~\ref{eq:collapsed}. Assume all of them directly sample from their respective conditional distributions for fairness consideration. We depict the convergence rate of a sampler with its spectral gap, which is defined in \ref{def:spectral-gap}.

\begin{definition}
\label{def:spectral-gap}
    For a Markov transition kernel $\mathbf{F}$, its \textbf{spectral gap} $Gap(\mathbf{F})$ is defined as:
    $$
    Gap(\mathbf{F}) = 1 - \lambda_1,
    $$
    where $\lambda_1$ is its second largest eigenvalue, also known as the spectral radius for reversible transition kernels (\cite{liu1994collapsed}).
\end{definition}

Roughly speaking, the spectral gap of a Markov transition kernel describes its convergence rate to the stationary distribution. To be more specific, a kernel $\mathbf{F}$ with a spectral gap $Gap(\mathbf{F})$ converges at a geometric rate in total variational distance (\cite{liu2001monte}):
$$
||\pi_t - \pi||_{TV} \leq (1-Gap(\mathbf{F}))^t ||\pi_0 - \pi||_{TV},
$$
where $\pi$ is the stationary distribution, and $\pi_t$ represents the law at step $t$. A forward derivation of the above result can be found in \cite{mcmc-tall-data}. It is easy to see that a chain with a larger spectral gap tends to converge faster.

\begin{theorem}\label{thm:rate}
    The spectral gaps of the three Gibbs samplers are ordered as:
    $$
    Gap(\mathbf{F}_c) \geq Gap(\mathbf{F}_m) \geq Gap(\mathbf{F}_g).
    $$
\end{theorem}
\begin{proof}[Proof of Theorem~\ref{thm:rate}]
    We follow the roadmap established in \cite{liu1994collapsed}. The three samplers' visiting schemes can be captured by the following diagram:
    \begin{equation}
        \begin{aligned}
            &\mathbf{F}_g: \bm{\theta} \rightarrow \mathbf{y}_m \rightarrow \{\mathbf{z}_m, \mathbf{z}_o\}\\
            &\mathbf{F}_m: \bm{\theta} \rightarrow \{\mathbf{z}_m,\mathbf{z}_o\}\\
            &\mathbf{F}_c: \bm{\theta} \rightarrow \mathbf{z}_o.
        \end{aligned}
    \end{equation}
    From Theorem 1 of \cite{liu1994collapsed}, it can be immediately derived that:
    $$
    ||\mathbf{F}_c|| \leq ||\mathbf{F}_m|| \leq ||\mathbf{F}_g||,
    $$
    where $||\mathbf{F}||$ represents the operator norm of a transition kernel $\mathbf{F}$. Since all three operators are reversible, their spectral radius equal to their operator norms. Because the spectral gap equals to one minus the spectral radius for reversible kernels (\cite{liu2001monte,liu1994collapsed}), the desired result can be derived immediately. $\blacksquare$
\end{proof}

Theorem~\ref{thm:rate} suggests that the proposed sampler converges at least as fast as the method proposed in \cite{yeh2012,Maarten2021}, whose theoretical convergence speed surpasses that of the classical Gibbs sampler. Therefore, we demonstrate that the proposed method has lower computational complexity. A faster convergence rate indicates that the proposed sampler can quickly mix with the target distribution and has a higher effective sample size (\cite{ess}).

\subsubsection{Computational Complexity Analysis}
\label{sec:computational-complexity}
We show that the proposed sampler is computationally faster than other competitive methods. The main computational bottleneck of numerical algorithms related to hidden Markov models (HMMs) involves the computation of forward or backward probabilities. All works related to hidden Markov models, including those dealing with missing observations (\cite{yeh2012,Maarten2021}), compute these probabilities recursively from $1$ to $T$ and have a time complexity of $O(nT)$, where $n$ is the sample size and $T$ is the sequence length (without loss of generality, we assume all sequences have length $T$).

However, Eq~\ref{eq:backward-prob} and Eq~\ref{eq:forward-prob} show that forward and backward probabilities only need to be computed recursively across all latent states corresponding to the observations in $\mathbf{y}_o$ if we integrate the missing observations and the latent states out. This approach only has a computational complexity of $O((1-p)nT)$, where $p$ is the missing rate defined in Eq~\ref{eq:missing-rate}. As we can see, when $p \to 0$, the time complexity approximates that of the algorithms presented in \cite{yeh2012,hidden-semi}, which represent the fully-observed scenario. However, when $p \to 1$, which indicates the presence of a large number of missing observations, the computational complexity decreases significantly. This fact suggests that the proposed algorithm is computationally faster than other competitors when the missing rate is high, making it particularly suitable for datasets with sequences that contain many missing entries.

Finally, we point out that although matrix exponential is needed in Eq~\ref{eq:backward-prob} and Eq~\ref{eq:forward-prob}, we can cache the multiplication results of the transition matrix $\mathbf{A}$ in advance and store them in a dictionary. With this optimization technique, the matrix $\mathbf{A}^k$ for $k=1, \ldots, T$ can be directly looked up from the dictionary, which has a computational complexity of $O(1)$. Therefore, the only time complexity that arises from evaluating transition matrix exponential comes from the precomputing and caching procedure, which has a time complexity of $O(T^2)$ in the worst case. Consequently, the overall computational complexity of updating $\mathbf{z}_o$ becomes $O(npT+T^2) \approx O(npT)$ when $n \gg T$. Thus, the computational complexity introduced by matrix multiplication is generally negligible. Additionally, under the scenario of a blockwise missing mechanism, the computational cost can be further reduced as long as the length of the missing block varies around a fixed length.

\section{Simulation Study}
\label{sec:sim}
In this section, we evaluate the performance of the proposed sampler through numerical simulations. We compare the performance of our sampler with three competitive methods: the EM algorithm (EM), the \textit{vanilla Gibbs} sampler (as described in \ref{eq:conditional}), and a Gibbs sampler that targets a distribution with $\mathbf{y}_m$ integrated out as described in \cite{yeh2012,Maarten2021} (whose scheme can be described by \ref{eq:ym-integrated}, abbreviated as \textit{partially-collapsed Gibbs}). We abbreviate the proposed method as \textit{collapsed Gibbs}.

We demonstrate the advantage of the proposed sampler via numerical simulations in three steps: First, we show that the proposed sampler has comparable performance in terms of estimation and prediction accuracy. Second, we show that the proposed sampler computes faster since it takes less time to generate a fixed number of samples. Third, we demonstrate that the proposed sampler is also computationally efficient in terms of ESS, that is, it has a larger ESS per second. Finally, we show that the proposed sampler is theoretically faster because its ESS per iteration is also larger than that of its competitors. In summary, the proposed method performs comparably with others in terms of estimation but is advantageous in terms of computational efficiency and sampling efficiency. All experiments are conducted on a Laptop equipped with the AMD Ryzen 7 5800H CPU and 32GB memory.

We consider a hidden Markov model with the following parameterization:
$$
\bm{\pi}=(0.6,0.3,0.1),
$$
$$
\mathbf{A}=
\begin{pmatrix}
0.6 & 0.3 & 0.1\\
0.1 & 0.6 & 0.3\\
0.3 & 0.1 & 0.6
\end{pmatrix},
$$
$$
\mathbf{B}=
\begin{pmatrix}
0.8 & 0.1 & 0.1\\
0.1 & 0.8 & 0.1\\
0.1 & 0.1 & 0.8
\end{pmatrix},
$$
As discussed in the last section, $\bm{\pi}$ stands for the initial distribution, $\mathbf{A}$ stands for transition matrix and $\mathbf{B}$ represents emission distribution.

\subsection{Simulation Study I: Random Missing}
\label{sec:sim-rs}
In this section, we conduct the simulation under the random missing mechanism. We generate $n=500$ sequences, each with length $T=20$. All observations are randomly dropped with a probability $p$. We set the probability $p$ to be $0, 0.1, 0.3, 0.5, 0.7, 0.9$. When the missing probability $p=0$, it corresponds to the scenario where the sequences are fully observed.

First, we compare the estimation accuracy. For parameters, the estimation accuracy is evaluated by the Mean Square Error (MSE) between the ground truth and the posterior mean. For latent states, the algorithms' performance is measured by the prediction accuracy on latent states $\mathbf{z}$ according to the majority vote summarized from the posterior. For the EM algorithm, we predict the hidden states by first estimating the parameters and then deriving the predictions via the Viterbi algorithm (\cite{forney1973viterbi}). We draw $5000$ samples from the posterior distribution and discard the first half as burn-in samples.

Table~\ref{tab:consolidated-metrics} shows that the estimation error and prediction accuracy of different samplers are comparable, and no significant difference in estimation accuracy between the samplers has been observed under different missing rates. This suggests that the proposed method is as effective as other methods in approximating the posterior.

\begin{table}
\caption{Estimation performance across different missing probabilities (under the random missing pattern) from posterior distributions generated by different samplers. All samplers generate distributions that have similar performance in posterior approximation at a fixed missing probability level. }
\label{tab:consolidated-metrics}
\begin{center}
\resizebox{\textwidth}{!}{%
\begin{tabular}{lllllll}
\toprule
  Missing Prob & Model & Latent State Prediction Accuracy & Initial Error & Transition Error & Emission Error \\
\midrule
0\% & Partially-collapsed Gibbs & 0.79 (0.01) & 0.0050 (0.00354) & 0.00063 (0.00060) & 0.0010 (0.00133) \\
     & Collapsed Gibbs & 0.79 (0.011) & 0.0050 (0.00352) & 0.00088 (0.00090) & 0.0012 (0.00139) \\
     & Vanilla Gibbs & 0.79 (0.01) & 0.0026 (0.00184) & 0.00065 (0.00099) & 0.0007 (0.00064) \\
     & EM & 0.76 (0.021) & 0.0062 (0.00421) & 0.00118 (0.00120) & 0.0015 (0.00150) \\
10\% & Partially-collapsed Gibbs  & 0.76 (0.0131)& 0.0080 (0.00568) & 0.00115 (0.00119) & 0.0015 (0.00219) \\
     & Collapsed Gibbs & 0.76 (0.023) & 0.0009 (0.00062) & 0.00035 (0.00049) & 0.0005 (0.00091) \\
     & Vanilla Gibbs & 0.77 (0.031)& 0.0077 (0.00544) & 0.00085 (0.00065) & 0.0016 (0.00157) \\
     & EM & 0.65 (0.030) & 0.0118 (0.00733) & 0.00120 (0.00140) & 0.0022 (0.00250) \\
30\% & Partially-collapsed Gibbs & 0.71 (0.0121)& 0.0014 (0.00098) & 0.00062 (0.00058) & 0.0008 (0.00071) \\
     & Collapsed Gibbs & 0.70 (0.021)& 0.0017 (0.00075) & 0.00031 (0.00025) & 0.0004 (0.00044) \\
     & Vanilla Gibbs & 0.70 (0.023)& 0.0006 (0.00042) & 0.00073 (0.00064) & 0.0010 (0.00103) \\
     & EM & 0.57 (0.025) & 0.0021 (0.00150) & 0.0010 (0.00090) & 0.0018 (0.00200) \\
50\% & Partially-collapsed Gibbs & 0.64 (0.0191)& 0.0013 (0.00092) & 0.00388 (0.00474) & 0.0020 (0.00161) \\
     & Collapsed Gibbs & 0.64 (0.0201)& 0.0018 (0.00125) & 0.00360 (0.00305) & 0.0031 (0.00315) \\
     & Vanilla Gibbs & 0.64 (0.0212)& 0.0137 (0.00969) & 0.00294 (0.00317) & 0.0029 (0.00260) \\
     & EM & 0.47 (0.015) & 0.0045 (0.00300) & 0.0040 (0.00400) & 0.0045 (0.00450) \\
70\% & Partially-collapsed Gibbs & 0.56 (0.022)& 0.0093 (0.00658) & 0.00399 (0.00396) & 0.0035 (0.00445) \\
     & Collapsed Gibbs & 0.55 (0.049)& 0.0197 (0.01393) & 0.01290 (0.01672) & 0.0110 (0.01841) \\
     & Vanilla Gibbs & 0.54 (0.02)& 0.0006 (0.00042) & 0.00244 (0.00209) & 0.0026 (0.00282) \\
     & EM & 0.45 (0.018) & 0.0205 (0.01512) & 0.0130 (0.01200) & 0.0115 (0.01150) \\
90\% & Partially-collapsed Gibbs & 0.40 (0.0231)& 0.0011 (0.00079) & 0.00681 (0.00793) & 0.0092 (0.01084) \\
     & Collapsed Gibbs & 0.38 (0.022)& 0.0154 (0.01089) & 0.02056 (0.01874) & 0.1226 (0.09503) \\
     & Vanilla Gibbs & 0.37 (0.0218)& 0.0140 (0.00987) & 0.07118 (0.08169) & 0.1536 (0.21586) \\
     & EM & 0.36 (0.030) & 0.0254 (0.02010) & 0.0300 (0.02950) & 0.1500 (0.14000) \\
\bottomrule
\end{tabular}
}
\end{center}
\end{table}

\begin{figure}
\label{fig:mar-time}
    \centering
    \includegraphics[width=0.5\linewidth]{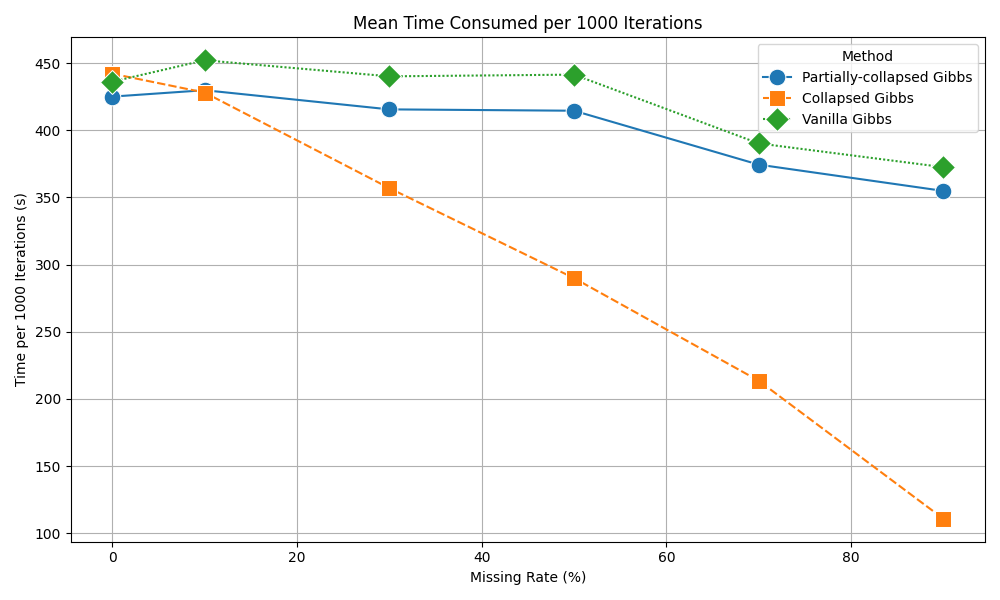}
    \caption{Time consumed per 1000 iterations of different samplers under different missing probabilities with the random missing mechanism. We only present the averaged time consumption because the computational complexity is deterministic (In fact, the standard deviation of computational time across different experiments is also very small, mainly caused by the jitter of the system). The line with circle markers represents the averaged consumed time of the partially collapsed Gibbs sampler and the line with diamond markers represents the consumed time of the vanilla Gibbs sampler. The average consumed time of the proposed collapsed sampler under different missing probabilities is specified by the line with a square markers. As the missing rate goes high, the proposed method becomes significantly faster than the competitive methods in terms of absolute time consumed per iteration under the random missing mechanism. Moreover, a linear decay in our method's computational time can be concluded from the figure, which verifies the results presented in complexity analysis.}
    \label{fig:enter-label}
\end{figure}

Based on the observation that all samplers have similar performance in terms of estimation, we argue that the advantage of the proposed sampler lies in its efficiency in posterior sampling. Specifically, it can produce more samples in a given time, achieves higher Effective Sample Size (ESS) per iteration, and also yields a larger ESS in a given period. Figure~\ref{fig:mar-time} first displays the results of the time each sampler takes to run 1000 iterations. All methods perform similarly when the missing probability is set to $0\%$, corresponding to the fully-observed case. However, as the missing probability increases, the proposed sampler begins to run faster than the competitive methods. When the missing probability reaches $90\%$, the proposed method is nearly four times faster than the competitive methods, demonstrating its advantage in terms of computational complexity. Moreover, the proposed method directly displays a pattern of linear decay in time complexity as the missing probability increases.

We next compare the samplers' theoretical sampling efficiency by examining the ESS produced per iteration. Given the numerous parameters in our model, we evaluate only the median of the ESS. As suggested in Table~\ref{tab:ess-per-iter}, the effective sample size generated by the proposed sampler consistently outperforms competitive methods when missing observations are involved. The results presented in Table~\ref{tab:ess-per-iter} numerically verify that the proposed sampler is theoretically more efficient than existing methods (in terms of convergence rate). Since the proposed method has a higher ESS per iteration and consumes less time per 1000 iterations compared to other algorithms, it consequently has a higher ESS per second, demonstrating its superior efficiency in posterior exploration. Combined with the running time results described in Figure~\ref{fig:mar-time}, we empirically demonstrate that the proposed method is fast at exploring the posterior both theoretically (measured by ESS per iteration) and computationally (measured by ESS per second and running time).

\begin{table}[ht]
\centering
\caption{Comparison of Gibbs sampling methods at missing probabilities in terms of ESS per iteration. The proposed method consistently produces a larger effective sample size per iteration when missing observations are involved.}
\label{tab:ess-per-iter}
\begin{tabular}{lcccc}
\toprule
Missing Prob & Vanilla Gibbs & Collapsed Gibbs & Partially-Collapsed Gibbs \\
\midrule
0\%   & 0.0067 (0.0005) & 0.0067 (0.0005) & \textbf{0.0072 (0.0004)} \\
10\%  & 0.0036 (0.0003) & \textbf{0.0038 (0.0003)} & 0.0034 (0.0002) \\
30\%  & 0.0019 (0.0002) & \textbf{0.0026 (0.0004)} & 0.0023 (0.0003) \\
50\%  & 0.0024 (0.0003) & \textbf{0.0036 (0.0005)} & 0.0026 (0.0003) \\
70\%  & 0.0016 (0.0002) & \textbf{0.0031 (0.0004)} & 0.0019 (0.0002) \\
90\%  & 0.0010 (0.0001) & \textbf{0.0047 (0.0006)} & 0.0008 (0.0001) \\
\bottomrule
\end{tabular}
\end{table}

\subsection{Simulation Study II: Blockwise Missing}
\label{sec: sim-bm}
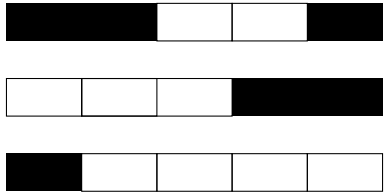
\begin{figure}[H]
\centering
\label{fig:blockwise-missing}
\begin{tikzpicture}
  \fill[black] (0, 0) rectangle (1, 0.5);
  \draw (1, 0) rectangle (1, 0.5);
  \fill[black] (1, 0) rectangle (2, 0.5);
  \draw (2, 0) rectangle (3, 0.5);
  \draw (3, 0) rectangle (4, 0.5);
  \fill[black] (4, 0) rectangle (5, 0.5);

  \draw (0, -1) rectangle (1, -0.5);
  \draw (1, -1) rectangle (2, -0.5);
  \draw (1, -1) rectangle (3, -0.5);
  \fill[black] (3, -1) rectangle (4, -0.5);
  \fill[black] (4, -1) rectangle (5, -0.5);

  \fill[black] (0, -2) rectangle (1, -1.5);
  \draw (1, -2) rectangle (2, -1.5);
  \draw (2, -2) rectangle (3, -1.5);
  \draw (3, -2) rectangle (4, -1.5);
  \draw (4, -2) rectangle (5, -1.5);
\end{tikzpicture}
\caption[Sequences with the blockwise missing pattern]{Sequences with the blockwise missing pattern. White blocks represent missing entries in the observed sequences }
\end{figure}

We continue our simulation study by considering the blockwise missing case. In the blockwise missing setting, instead of setting the missing entries randomly, we designate a continuous block in the sequence to be missing. Figure~\ref{fig:blockwise-missing} provides a graphical illustration of the blockwise missing pattern. The black cells represent the observed entries, while the white continuous blocks stand for the missing blocks in the observed sequences.

\begin{figure}
    \centering
    \includegraphics[width=0.5\linewidth]{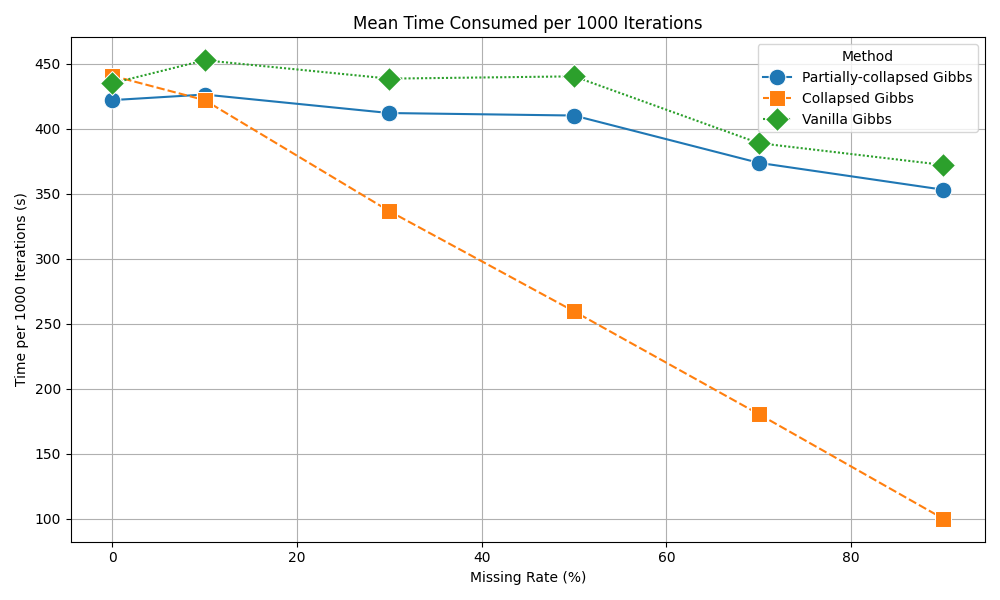}
    \caption{Comparison of average time consumed per 1000 iterations over different samplers with the block missing mechanism. Since the computational complexity is deterministic and there is little uncertainty, we only present the mean (In fact, the true standard deviation is also rather small, mainly caused by system's jitter). The line with circle markers represents the averaged consumed time of the partially collapsed Gibbs sampler and the line with diamond markers represents the consumed time of the vanilla Gibbs sampler. The average consumed time of the proposed collapsed sampler under different missing probabilities is specified by the line with a square markers. The proposed method is significantly faster than any other existing methods when the missing probability is high with the block missing mechanism. Moreover, a linear decay in our method's computational time can be concluded from the figure, which verifies the results presented in complexity analysis.}
    \label{fig:Block-time}
\end{figure}

To generate incomplete sequences with blockwise missing, we proceed as follows: for each sequence, a random index set $\{i, i+1, \ldots, i+j\}$ is selected, where $j \approx T \times p$. The observations within these indices are set to be missing. The numerical experiments follow exactly the same procedures discussed in the previous section. Table~\ref{tab:performance-across-missing-probs-block} records the estimation accuracy of different algorithms under various levels of missing data. It can be clearly seen that the proposed algorithm has similar estimation accuracy compared to existing methods. Figure~\ref{fig:Block-time} provides a direct comparison of the time consumed per 1000 iterations and suggests that the proposed sampler consistently consumes less time than existing methods when the missing probability is high, demonstrating its advantage in computational efficiency. Moreover, the proposed method directly displays a pattern of linear decay in time complexity as the missing probability increases. Similarly, the median of the ESS per iteration is compared in Table~\ref{tab:block-ess-per-iter}, demonstrating the theoretical sampling efficiency. To summarize, we empirically verify that the proposed method is faster than existing algorithms both computationally and theoretically in posterior exploration when tackling datasets with blockwise missing.

\begin{table}[ht]
\centering
\caption{Estimation performance across different missing probabilities (under the blockwise missing pattern) from posterior distributions generated by different models. All models generate distributions that have similar performance in posterior approximation at a fixed missing probability level.}
\label{tab:performance-across-missing-probs-block}
\resizebox{\textwidth}{!}{%
\begin{tabular}{llllll}
\toprule
Missing Prob & Model & Latent State Prediction Accuracy & Initial Error & Transition Error & Emission Error \\
\midrule
0\% & Vanilla Gibbs & 0.78 (0.023) & 0.0028 (0.0020) & 0.0004 (0.0005) & 0.0007 (0.0010) \\
    & Collapsed Gibbs & 0.78 (0.020) & 0.0083 (0.0059) & 0.0015 (0.0014) & 0.0029 (0.0035) \\
    & Partially-Collapsed Gibbs & 0.78 (0.022) & 0.0038 (0.0027) & 0.0008 (0.0010) & 0.0010 (0.0013) \\
    & EM & 0.76 (0.030) & 0.0120 (0.0090) & 0.0025 (0.0026) & 0.0040 (0.0041) \\
10\% & Vanilla Gibbs & 0.76 (0.018) & 0.0002 (0.0002) & 0.0016 (0.0023) & 0.0018 (0.0019) \\
    & Collapsed Gibbs & 0.74 (0.019) & 0.0059 (0.0042) & 0.0014 (0.0019) & 0.0010 (0.0016) \\
    & Partially-Collapsed Gibbs & 0.76 (0.022) & 0.0024 (0.0017) & 0.0012 (0.0011) & 0.0008 (0.0009) \\
    & EM & 0.72 (0.025) & 0.0085 (0.0060) & 0.0028 (0.0029) & 0.0035 (0.0036) \\
30\% & Vanilla Gibbs & 0.68 (0.015) & 0.0043 (0.0031) & 0.0008 (0.0011) & 0.0014 (0.0013) \\
    & Collapsed Gibbs & 0.65 (0.030) & 0.0007 (0.0005) & 0.0012 (0.0014) & 0.0012 (0.0012) \\
    & Partially-Collapsed Gibbs & 0.67 (0.014) & 0.0003 (0.0002) & 0.0015 (0.0024) & 0.0021 (0.0022) \\
    & EM & 0.57 (0.035) & 0.0020 (0.0018) & 0.0023 (0.0024) & 0.0030 (0.0031) \\
50\% & Vanilla Gibbs & 0.59 (0.022) & 0.0021 (0.0015) & 0.0033 (0.0039) & 0.0032 (0.0037) \\
    & Collapsed Gibbs & 0.56 (0.025) & 0.0020 (0.0014) & 0.0031 (0.0035) & 0.0026 (0.0033) \\
    & Partially-Collapsed Gibbs & 0.58 (0.013) & 0.0014 (0.0010) & 0.0017 (0.0012) & 0.0019 (0.0019) \\
    & EM & 0.47 (0.028) & 0.0050 (0.0040) & 0.0045 (0.0046) & 0.0048 (0.0049) \\
70\% & Vanilla Gibbs & 0.49 (0.027) & 0.0034 (0.0024) & 0.0024 (0.0039) & 0.0035 (0.0046) \\
    & Collapsed Gibbs & 0.47 (0.014) & 0.0143 (0.0101) & 0.0049 (0.0036) & 0.0047 (0.0039) \\
    & Partially-Collapsed Gibbs & 0.50 (0.020) & 0.0039 (0.0028) & 0.0031 (0.0028) & 0.0020 (0.0023) \\
    & EM & 0.45 (0.032) & 0.0200 (0.0150) & 0.0058 (0.0050) & 0.0060 (0.0055) \\
90\% & Vanilla Gibbs & 0.38 (0.029) & 0.0094 (0.0067) & 0.1425 (0.1414) & 0.0606 (0.0666) \\
    & Collapsed Gibbs & 0.37 (0.021) & 0.0052 (0.0037) & 0.0095 (0.0099) & 0.0152 (0.0160) \\
    & Partially-Collapsed Gibbs & 0.37 (0.016) & 0.0167 (0.0118) & 0.0146 (0.0159) & 0.0219 (0.0239) \\
    & EM & 0.35 (0.035) & 0.0250 (0.0200) & 0.0150 (0.0140) & 0.0200 (0.0190) \\
\bottomrule
\end{tabular}
}
\end{table}

\begin{table}[ht]
\centering
\caption{Comparison of Gibbs sampling methods at missing probabilities in terms of the median of ESS per iteration (with blockwise missing mechanism). The proposed method consistently produces a larger effective sample size per iteration when blockwise missing observations are involved.}
\label{tab:block-ess-per-iter}
\begin{tabular}{lcccc}
\toprule
Missing Prob & Vanilla Gibbs & Collapsed Gibbs & Partially-Collapsed Gibbs \\
\midrule
0\%   & \textbf{0.0046 (0.0003)} & 0.0039 (0.0002) & 0.0044 (0.0003) \\
10\%  & 0.0029 (0.0004) & \textbf{0.0049 (0.0003)} & 0.0027 (0.0002) \\
30\%  & 0.0025 (0.0003) & \textbf{0.0037 (0.0002)} & 0.0036 (0.0002) \\
50\%  & 0.0010 (0.0001) & \textbf{0.0050 (0.0004)} & 0.0016 (0.0001) \\
70\%  & 0.0013 (0.0001) & \textbf{0.0025 (0.0002)} & 0.0011 (0.0001) \\
90\%  & 0.0011 (0.0001) & \textbf{0.0107 (0.0005)} & 0.0058 (0.0003) \\
\bottomrule
\end{tabular}
\end{table}

\section{Real Data Analysis}
\label{sec: rda}
In the real data analysis section, we employ our framework to analyze two distinct datasets. Initially, we apply the proposed algorithm to the schizophrenia dataset (\cite{hedeker1997}) and the CLEAR dataset (\cite{huang2019}) to compare the sampling efficiency between the proposed method and existing algorithms. All experiments are conducted on a Laptop equipped with the AMD Ryzen 7 5800H CPU and 32GB memory.

\subsection{Real Data Analysis I: Schizophrenia Study}
\label{sec: rda-ss}
In this section, we fit a Bayesian hidden Markov model to the schizophrenia dataset, a public disease progression dataset released by the National Institute of Mental Health, and compare the sampling efficiency across various methods.

The schizophrenia dataset (\cite{hedeker1997}) comprises incomplete observational sequences from $437$ distinct patients diagnosed with schizophrenia. Of these patients, $108$ were prescribed a placebo, while the remaining $329$ received medication. Their health status was monitored over six consecutive weeks, during which they reported their mental status daily. Mental status is a categorical variable categorized into four levels: severe, moderate, mild, or normal. The dataset was initially studied in 1997, focusing on the application of the pattern mixture model. More recent studies, such as \cite{yeh2012,Maarten2021}, have employed a Maximum Likelihood Estimation (MLE) approach to model this dataset as a Hidden Markov Model (HMM).

Given that patients suffering from schizophrenia may report their status inaccurately, it is both natural and advantageous to model this process using a hidden Markov model. Indeed, the approach of treating the true disease status as an unknown hidden variable and the observed status as a variable sampled from an emission process is well-established in the literature on medical record analyses (\cite{Altman2005}).

In the schizophrenia dataset, approximately $50\%$ of the entries are missing. Specifically, observations from weeks $2$, $4$, and $5$ have over $90\%$ of their entries missing. Concurrently, about $30\%$ of the entries in weeks $3$ and $6$ are missing. This pattern of missingness suggests that the omissions may be attributable to the study's design, as they are closely related to specific dates, and thus can be considered ignorable.

To analyze the data, we adopt the following approach: We train two distinct hidden Markov models, one for patients prescribed a placebo and another for those administered the medication under investigation. For each model, we initialize the parameters from a uniform distribution and let the Gibbs sampler run for $5000$ steps. The first $2500$ samples are discarded as burn-in, and the remaining $2500$ samples are used as posterior draws. The estimated posterior mean and std produced by the proposed method are detailed in Table~\ref{table:real data1} for reference. Each experiment is replicated ten times. A comprehensive description of the estimated posterior is provided in the supplement. The solid red lines in these figures represent the estimated posterior means.

Apart from posterior sampling, we place our main focus on sampling efficiency. Since ground truth parameters are not available, we assess the performance of the estimation by randomly masking some existing observations and let the models to predict them. The prediction accuracy on the missing observations is then compared to the ground truth and evaluated to quantitatively assess the estimation accuracy of the posteriors produced by different samplers. We name this metric as the \textit{Cross-Validated Prediction Accuracy} in the report. Additionally, similar to Sections~\ref{sec:sim-rs} and \ref{sec: sim-bm}, we report the comparison between each sampler on average time complexity, convergence rate (measured by ESS per iteration), and overall sampling efficiency (measured by the number of ESS over a period of time). Table~\ref{tab:schizrep-combined} presents the respective results evaluated on the treatment group and the control group across various metrics. In summary, all algorithms achieve similar results in posterior prediction, while the proposed collapsed sampler is computationally faster and produces a larger ESS per iteration, thus yielding a higher ESS in a given period of time averaged over 10 experiments. Moreover, it further verifies the proposed sampler's advantage in terms of time complexity under scenarios where the missing probability is high. Therefore, results summarized from Table~\ref{tab:schizrep-combined} suggest that the proposed sampler exhibits advantages in terms of lower computational complexity and better convergence rate, as argued in previous sections.

\begin{table}[ht]
\centering
\caption{Comparison of Gibbs sampling methods on the groups of patients prescribed with drugs and placebos. Empirical evaluations suggest that the proposed method is more efficient in posterior exploration computationally and theoretically because it takes fewer time (for computational time, we only report the mean because the only uncertainty comes from the computer's random jitter) for each iteration and the effective sample produced per iteration is higher than other competitive algorithms.}
\label{tab:schizrep-combined}
\resizebox{\textwidth}{!}{%
\begin{tabular}{llcccc}
\toprule
& & \multicolumn{4}{c}{Metrics} \\
\cmidrule{3-6}
Group & Sampler & Time per 1000 iterations & Median ESS per iteration & Median ESS per second & Cross-Validated Prediction Accuracy \\
\midrule
\multirow{3}{*}{Treatment Group}
& Vanilla Gibbs & 126.95  & 0.0031 (0.0003) & 0.024 & 45.21\% (2.3\%) \\
& Collapsed Gibbs & \textbf{82.99} & \textbf{0.0036} (0.0005) & \textbf{0.044} & 44.57\% (3.1\%) \\
& Partially-Collapsed Gibbs & 110.51  & 0.0011 (0.0008) & 0.010 & \textbf{45.34\%} (1.7\%) \\
\midrule
\multirow{3}{*}{Control Group}
& Vanilla Gibbs & 127.59  & 0.0022 (0.0011) & 0.017 & \textbf{47.88\%} (3.2\%) \\
& Collapsed Gibbs & \textbf{82.07}  & \textbf{0.016} (0.0036) & \textbf{0.200} & 47.56\% (3.0\%) \\
& Partially-Collapsed Gibbs & 109.77  & 0.014 (0.0007) & 0.130 & 47.20\% (3.2\%) \\
\bottomrule
\end{tabular}
}
\end{table}

\begin{table}
\centering
\caption{Estimated posterior mean (Std) in the Schizophrenia dataset study}
\label{table:real data1}
\resizebox{\textwidth}{!}{%
\begin{tabular}{@{}llcccccccc@{}}
\toprule
& & \multicolumn{4}{c}{\textbf{Placebo Group}} & \multicolumn{4}{c}{\textbf{Treatment Group}} \\ \cmidrule(lr){3-6} \cmidrule(lr){7-10}
& & \textbf{Normal} & \textbf{Mild} & \textbf{Moderate} & \textbf{Severe} & \textbf{Normal} & \textbf{Mild} & \textbf{Moderate} & \textbf{Severe} \\ \midrule
\multirow{1}{*}{\textbf{Initial Distribution}} & \textbf{} & 0.01($10^{-4}$) & 0.10(0.001) & 0.33(0.004) & 0.56(0.003) & 0.01($10^{-5}$) & 0.09($10^{-4}$) & 0.26(0.0020) & 0.62(0.0019) \\ \midrule
\multirow{4}{*}{\textbf{Transition Matrix}} & \textbf{Normal} & 0.69(0.019) & 0.12(0.011) & 0.10(0.008) & 0.09(0.007) & 0.94(0.001) & 0.05(0.001) & 0.01(0.000) & 0.01(0.000) \\
& \textbf{Mild} & 0.04(0.001) & 0.90(0.002) & 0.05(0.001) & 0.02(0.000) & 0.20(0.001) & 0.78(0.001) & 0.02(0.000) & 0.01(0.000) \\
& \textbf{Moderate} & 0.02(0.001) & 0.19(0.003) & 0.73(0.005) & 0.07(0.001) & 0.05(0.000) & 0.26(0.001) & 0.66(0.002) & 0.03(0.000) \\
& \textbf{Severe} & 0.01(0.000) & 0.03(0.000) & 0.08(0.001) & 0.87(0.001) & 0.03(0.000) & 0.11(0.001) & 0.30(0.002) & 0.56(0.001) \\ \midrule
\multirow{4}{*}{\textbf{Emission Matrix}} & \textbf{Normal} & 0.62(0.026) & 0.13(0.014) & 0.14(0.014) & 0.11(0.010) & 0.86(0.005) & 0.11(0.005) & 0.01(0.000) & 0.01(0.000) \\
& \textbf{Mild} & 0.03(0.001) & 0.86(0.004) & 0.08(0.004) & 0.03(0.001) & 0.01(0.000) & 0.93(0.001) & 0.05(0.001) & 0.01(0.000) \\
& \textbf{Moderate} & 0.01(0.000) & 0.10(0.004) & 0.82(0.005) & 0.06(0.002) & 0.01(0.000) & 0.19(0.003) & 0.75(0.003) & 0.06(0.002) \\
& \textbf{Severe} & 0.01(0.000) & 0.02(0.000) & 0.06(0.001) & 0.91(0.001) & 0.00(0.000) & 0.01(0.000) & 0.07(0.002) & 0.91(0.002) \\ \bottomrule
\end{tabular}
}
\end{table}

\subsection{Real Data Analysis II: MRSA Infection study}
\label{sec: rda-mis}
In this section, we apply the proposed algorithm to the CLEAR dataset (\cite{huang2019}) for comprehensive medical record analysis and compare its sampling efficiency with existing algorithms. The CLEAR dataset tracks patients from various groups regarding their Methicillin-resistant Staphylococcus Aureus (MRSA) infection and colonization status.

MRSA is a type of dangerous bacterium that may be found in human bodies and is generally difficult to eradicate (\cite{MRSAReview}). A serious MRSA infection can lead to severe sepsis and even be life-threatening. In this study, patients previously infected with MRSA were recruited at the beginning of the experiment and were divided into a treatment group and a control group. A repeated decolonization protocol was then applied to patients in the treatment group. Patients from both groups received education on general hygiene and environmental cleaning. The treatment group comprises $1058$ patients, while the control group includes $1063$ patients.

Samples were collected from the patients' throats on days $0, 1, 3$, and $6$, with test results provided to determine MRSA colonization in each patient's sample. Consequently, observations on days $2, 4, 5$ are considered missing data. Additionally, observations on days $0, 1, 3$, and $6$ also exhibit many missing entries. In the control group, approximately $20\%$ of observations on days $1$ and $3$ are missing, while $33\%$ of observations on day $6$ are missing. The overall missing rate is around $55\%$. In the treatment group, approximately $30\%$ of observations on days $1, 3$, and $6$ are missing, with an overall missing rate of about $57\%$. Generally, the sequences exhibit a blockwise missing structure, supplemented by some random missing.

Building on the treatment described in the previous section, we model the test results as inaccurate observations of the true status of MRSA colonization, which is an unobserved latent state. This approach is valid because MRSA is typically difficult to eradicate, and a negative test result may not necessarily indicate complete eradication (\cite{MRSAReview}). Two distinct models are trained separately for the treatment and control groups. We generate $5000$ samples using MCMC and retain the last $2500$ samples as posterior draws. Similar to the settings in Section~\ref{sec: rda-mis}, we compare the performance of different samplers in terms of average prediction accuracy, time complexity, and convergence rate over eight replicates.

The estimated posterior mean and standard deviation are recorded in Table~\ref{table:real_data2}. This table presents the posterior mean and variance for each parameter estimated using our framework. For a more detailed visualization of the posterior distribution derived from the MCMC simulation, we refer the readers to the supplement. The posterior distribution of the transition matrix suggests that the treatment is more effective in eradicating MRSA infections. This is evidenced by the fact that patients in the treatment group generally exhibit significantly higher probabilities of transitioning from infection to cure, given that they are infected with MRSA.

Comparisons of the respective sampling efficiencies in the treatment and control groups can be found in Table~\ref{tab:clear-combined}. The results in this table reveal that all samplers achieve similar prediction scores. However, the proposed sampler is more efficient in posterior exploration, as it requires less time to run 1000 iterations and its Effective Sample Size (ESS) per iteration is significantly higher than that of competitive methods, resulting in a higher ESS per second. These results validate our conclusions established in Section~\ref{sec:complexity-analysis}, that the proposed sampler enjoys lower computational complexity and better convergence rate.

\begin{table}[ht]
\centering
\caption{Comparison of Gibbs sampling methods in the CLEAR dataset for treatment and control groups. Empirical evaluations suggest that the proposed method is more efficient in posterior exploration computationally and theoretically because it takes fewer time (for computational time, we only report the mean because the only uncertainty comes from the computer's random jitter) for each iteration and the effective sample produced per iteration is higher than other competitive algorithms.}
\label{tab:clear-combined}
\resizebox{\textwidth}{!}{%
\begin{tabular}{llcccc}
\toprule
& & \multicolumn{4}{c}{Metrics} \\
\cmidrule{3-6}
Group & Sampler & Time per 1000 iterations & Median ESS per iteration & Median ESS per second & Cross-validated Prediction Accuracy \\
\midrule
\multirow{3}{*}{Treatment Group}
& Vanilla Gibbs & 343.24  & 0.00017 (0.00012) & 0.00049 & 88.21\% (1.2\%) \\
& Collapsed Gibbs & \textbf{232.64}  & \textbf{0.0026} (0.0003) & \textbf{0.011} & 87.57\% (1.2\%) \\
& Partially-Collapsed Gibbs & 397.91  & 0.0017 (0.0003) & 0.0043 & \textbf{88.89\%} (0.9\%) \\
\midrule
\multirow{3}{*}{Control Group}
& Vanilla Gibbs & 387.92  & 0.00037 (0.0003) & 0.00095 & 82.01\% (0.9\%) \\
& Collapsed Gibbs & \textbf{287.85}  & \textbf{0.0019} (0.0004) & \textbf{0.00667} & \textbf{82.97\%} (1.3\%) \\
& Partially-Collapsed Gibbs & 320.83  & 0.0015 (0.0002) & 0.00467 & 80.13\% (1.8\%) \\
\bottomrule
\end{tabular}
}
\end{table}

\begin{table}
\centering
\caption{Estimated posterior mean (Std) in the MRSA infection study}
\label{table:real_data2}
\resizebox{\textwidth}{!}{%
\begin{tabular}{@{}lclclcclc@{}}
\toprule
& \multicolumn{1}{c}{\textbf{Initial Distribution}} & & \multicolumn{2}{c}{\textbf{Transition Matrix}} & & \multicolumn{3}{c}{\textbf{Emission Matrix}} \\ \cmidrule(lr){2-2} \cmidrule(lr){4-5} \cmidrule(lr){7-9}
\textbf{States} & \textbf{} & & \textbf{Cure} & \textbf{Infection} & & \textbf{Cure} & \textbf{Infection} \\ \midrule
Cure (Control Group) & 0.71 (0.028) & & 0.99(0.006) & 0.01(0.006) & & 0.94(0.008) & 0.06(0.008) \\
Infection (Control Group) & 0.29 (0.028) & & 0.11(0.032) & 0.89(0.032) & & 0.35(0.050) & 0.65(0.050) \\
Cure (Treatment Group) & 0.69 (0.035) & & 0.98(0.024) & 0.03(0.024) & & 0.99(0.011) & 0.01(0.011) \\
Infection (Treatment Group) & 0.31 (0.035) & & 0.29(0.004) & 0.72(0.004) & & 0.30(0.045) & 0.71(0.045) \\ \bottomrule
\end{tabular}
}
\end{table}

\section{Conclusions and Discussions}
In this paper, we propose a novel Gibbs sampling algorithm for a Bayesian hidden Markov model with missing data. The proposed algorithm samples from a collapsed distribution, with latent states corresponding to the missing observations integrated out. This approach reduces the computational complexity of latent state sampling and yields a better convergence rate, particularly when the probability of missing data is high. These advantageous properties are verified both theoretically and through empirical evidence, using simulations and real data analysis. Empirical evaluations on synthesized and real-world datasets demonstrate the computational advantage of the proposed algorithm in terms of ESS per iteration and ESS per second, compared with existing algorithms. Therefore, we conclude that by integrating out the irrelevant data, the proposed algorithm achieves superior computational and theoretical efficiency compared to existing algorithms.
\bigskip
\begin{center}
{\large\bf SUPPLEMENTARY MATERIAL}
\end{center}

\begin{description}
\item[Software:] Our code is publicly available at https://github.com/lidongrong/PHMM. Python package will also be released in the future.

\item In the supplement, we provide more numerical results on real-data applications, with detailed graphical descriptions of the posterior density provided.

\end{description}

\bibliographystyle{chicago}
\bibliography{Bibliography-MM-MC}
\end{document}